\documentclass[10pt,twocolumn,letterpaper]{article}

\usepackage{cvpr}      
\definecolor{cvprblue}{rgb}{0.21,0.49,0.74}
\usepackage[pagebackref,breaklinks,colorlinks,allcolors=cvprblue]{hyperref}

\def\paperID{*****} 
\def\confName{CVPR}
\def\confYear{2026}

\title{Autonomous UAV Navigation for Individual Wildlife Re-Identification}

\author{Claire Sun\\
The Ohio State University\\
{\tt\small sun.3903@osu.edu}
\and
Tanya Berger-Wolf \\
The Ohio State University
\and
Jenna Kline \\
The Ohio State University
}

\begin{document}
\maketitle
\begin{abstract}
Reliable individual re-identification (re-ID) of wildlife is essential for population monitoring, behavioral tracking, and conservation policy evaluation, yet large-scale data collection remains labor-intensive, relying on manual efforts by ecologists or citizen scientists. 
We propose an autonomous drone navigation system that actively optimizes image capture for downstream re-ID, moving beyond passive aerial sensing. 
The system combines YOLOv11 object detection with a DINOv2-based pose classifier to guide real-time flight decisions: detecting animals, orienting to expose the lateral flank (the surface of interest for pattern-based re-ID), and approaching until the subject meets a minimum bounding-box threshold. 
Unlike prior drone systems that optimize for group-level behavioral video, ours targets the specific image-quality requirements of individual-identification models. 
We demonstrate feasibility through a case study on zebra using footage collected in Kenya, and show the approach generalizes to other species with diagnostic surface patterns, including giraffes, tigers, and elephants. 
Our work establishes a framework for task-aware embodied AI for ecological data collection, in which downstream re-ID requirements drive real-time perception and control.
    
\end{abstract}

\section{Introduction}
Accurate population counts and longitudinal tracking of individual animals are foundational to conservation science. They underpin management decisions for endangered species and enable rigorous evaluation of protection policies. The Great Grevy's Rally, a citizen-science census of Grevy's zebra (\textit{Equus grevyi}, fewer than 3,000 remaining in the wild), illustrates both the value and the cost of this work: mobilizing hundreds of volunteers for a single multi-day survey \cite{grevyszebratrust2024annual}.

Individual re-identification (re-ID) of wildlife from natural markings, pioneered by systems such as HotSpotter \cite{crall2013hotspotter} and now extended by deep learning approaches \cite{Schneider2019Past, cermak2024Wildlife, Li2025Adaptive, Nepovinnykh2024Species}, has significantly reduced the need for invasive tagging. 
Platforms like Wildbook aggregate citizen-science images to track individuals across time and geography \cite{bergerwolf2017wildbook}.
However, the data collection bottleneck remains: images must be manually gathered, and their quality is highly variable. 
An image is only useful for re-ID if it captures the correct surface-of-interest (e.g., the lateral flank for zebras) at sufficient resolution.
These conditions are rarely guaranteed in opportunistic photography.

Unmanned aerial vehicles (UAVs) have transformed wildlife monitoring by enabling rapid, large-area surveys with reduced disturbance \cite{Corcoran2021Automated}. 
Recent work has moved from manual to autonomous UAV operation for behavioral studies \cite{Kline2026WildWing, meier2024wildbridge, Kline2025Studying}
demonstrating that integrating ecological knowledge into navigation policies improves data quality \cite{Kline2024Integrating}. 
However, these systems optimize for group-level behavioral video by maintaining altitude and herd coverage, rather than for individual identification, which demands fundamentally different image properties.

This paper addresses that gap with three contributions. (1) A task-driven navigation protocol (detect, orient, approach, and capture, or DOAC) that couples pose classification and flight control to maximize re-ID-quality image yield. (2) A DINOv2-based lateral pose classifier for zebra flank visibility, deployable on edge hardware. (3) Feasibility validation on real Kenya UAV footage, with a path to generalization across species with diagnostic surface patterns.

\begin{figure*}
    \centering
    \includegraphics[width=1\linewidth]{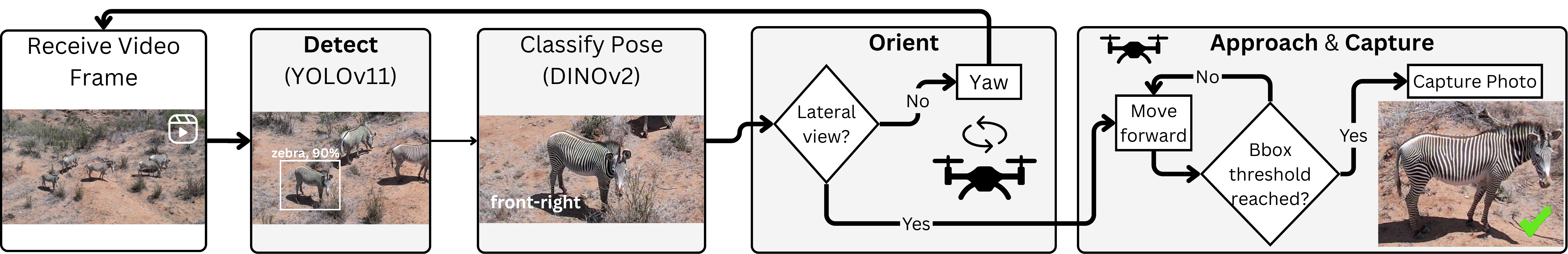}
    \caption{Navigation pipeline for individual zebra identification employing the detect-orient-approach-capture (DOAC) approach. The zebra is detected using YOLOv11, then its pose is classified using DINOv2. If a lateral view is detected, the drone moves forward until the bounding box (bbox) threshold is reached. If there is not a lateral view, such as the front-right view shown, the drone yaws left or right until a lateral view is captured, or the timeout condition is reached.}
    \label{fig: pipeline}
\end{figure*}

\section{Related Work}
\paragraph{Animal re-identification.} 
Re-identification of individual animals using computer vision has been demonstrated for zebras, tigers, elephants, giraffes, and many other species with distinctive coat patterns or morphologies \cite{Schneider2019Past}. 
Wildbook and ElephantBook provide platforms for managing re-ID at scale \cite{bergerwolf2017wildbook, Kulits2021ElephantBook}. 
Recent work has pushed toward species-agnostic re-ID. 
Čermák et al. introduce WildlifeDatasets and MegaDescriptor, a foundation model for multi-species re-ID \cite{cermak2024Wildlife}, 
while Li et al. propose a high-frequency transformer for unified multi-species identification \cite{Li2025Adaptive}. 
Nepovinnykh et al. combine learnable local features with aggregation for robust pattern matching across small datasets \cite{Nepovinnykh2024Species}. 
A consistent theme across these methods is their sensitivity to image quality, as accurate identification requires a well-defined pose and sufficient bounding-box resolution.

\paragraph{Drone-based wildlife monitoring.} 
UAVs have become standard tools for detection-focused wildlife surveys \cite{Corcoran2021Automated}. 
BuckTales introduces a multi-UAV re-ID and tracking dataset for wild blackbuck antelope \cite{Naik2024Bucktales} 
The In Situ Kenyan Animal Behavior Recognition (KABR) dataset provides a large collection of video clips of labeled zebra and giraffe behaviors \cite{Kholiavchenko2024KABR}.
Andrew et al. equip a UAV with onboard deep inference for biometric identification of individual Friesian cattle using a detector, an exploratory navigation network, and an identification network \cite{Andrew2019Deep}. 
Our work differs from this foundational work in targeting flank-based morphologically distinctive re-ID, explicitly formalizing pose as a navigation objective, and using a modern DINOv2-based classifier. Rolland et al. formalize surface-of-interest (SoI) coverage for drone swarms \cite{Rolland2024swarm}, and Kline et al. identify SoI as a critical planning variable for AI-driven aerial ecology missions \cite{Kline2025Studying}.
No existing system operationalizes SoI at the individual animal level in real time. Our drone navigation protocol does exactly this by actively seeking the lateral-flank SoI for each detected individual, closing the loop between mission-planning concepts and autonomous execution.

\paragraph{Autonomous UAV navigation for ecology.}
WildBridge \cite{meier2024wildbridge} demonstrates accurate animal localization from commercial UAVs but treats image capture as passive. Thus, optimizing the viewpoint for identification quality remains unaddressed.
WildWing \cite{Kline2026WildWing} and the imageomics framework of Kline et al. \cite{Kline2024Integrating} demonstrate autonomous navigation policies derived from expert pilot telemetry, but optimized for group behavioral video, not individual identification missions. 
Grushchak et al. present decentralized multi-drone herd coverage \cite{Grushchak2023Decentralized}. 
These systems treat data quality as a function of altitude, distance, and angle to the group.
In contrast, our system defines quality at the individual level through pose and resolution constraints, which is a necessary shift for re-ID.

\section{Method}

\subsection{Problem Formulation}
An image is re-ID suitable for a given individual if two conditions hold. One, the surface of interest is visible. 
For zebras, the lateral flank bears the unique stripe pattern. 
Two, the bounding box occupies sufficient image area for pattern matching. We use a minimum threshold of 500×500 pixels, following standard re-ID practice \cite{Kline2025Studying}.
Our navigation protocol is designed to maximize the yield of images satisfying both conditions.

The choice of lateral flank over top-down view is deliberate. While nadir (top-down) imagery is standard for population counting, individual re-ID relies on lateral flank patterns.
This is the view from which expert field researchers identify animals, and from which the majority of curated re-ID training data (camera traps, ground-level photography) has been collected \cite{Schneider2019Past, bergerwolf2017wildbook}. 
Targeting the lateral SoI allows our system to directly leverage existing annotated datasets and pretrained re-ID models, rather than requiring new data-collection paradigms.
However, if future models are developed to conduct individual identification from the nadir view, this framework can be easily adapted to accommodate.

\subsection{Navigation Pipeline}
The system operates in four sequential stages, executed in a real-time loop on onboard compute, illustrated in Fig. \ref{fig: pipeline}.

\begin{enumerate}
    \item \textbf{Detect.} A YOLOv11 model \cite{khanam2024yolov11} fine-tuned on aerial zebra imagery processes each video frame. Detected bounding boxes are ranked by confidence. The largest (nearest) detection is selected as the navigation target when multiple individuals are present.
    \item \textbf{Classify pose.} The detected crop is passed to a DINOv2-based pose classifier that predicts one of 8 discrete orientations arranged radially around the animal: front, front-left, left, back-left, back, back-right, right, and front-right, illustrated in Fig. \ref{fig:dataset}. The classifier uses a frozen DINOv2-small (ViT-S/14, 384-dim) \cite{oquab2023dinov2} backbone with a trainable MLP head (LayerNorm → 256 → 128 → 8), trained with label-aware horizontal flip augmentation that swaps mirrored class pairs (e.g., left $\leftrightarrow$ right). Poses in the left or right classes advance the pipeline as lateral SoI views. All others trigger the orient maneuver.
    \item \textbf{Orient.} If the pose is non-lateral, the drone rotates in place by incrementally adjusting the yaw until a lateral pose is detected or a timeout is reached, at which point the system searches for another individual.
    \item \textbf{Approach and capture.} Once a lateral pose is confirmed, the drone approaches slowly until the detected bounding box exceeds the 500×500 pixel threshold. At that point, image capture is triggered, and GPS metadata is logged.
\end{enumerate}

This detect–orient–approach–capture (\textbf{DOAC}) loop runs continuously, enabling the drone to respond to animal movement and recover from occlusion.

\subsection{Models and Training}

\begin{figure}
    \centering
    \includegraphics[width=1\linewidth]{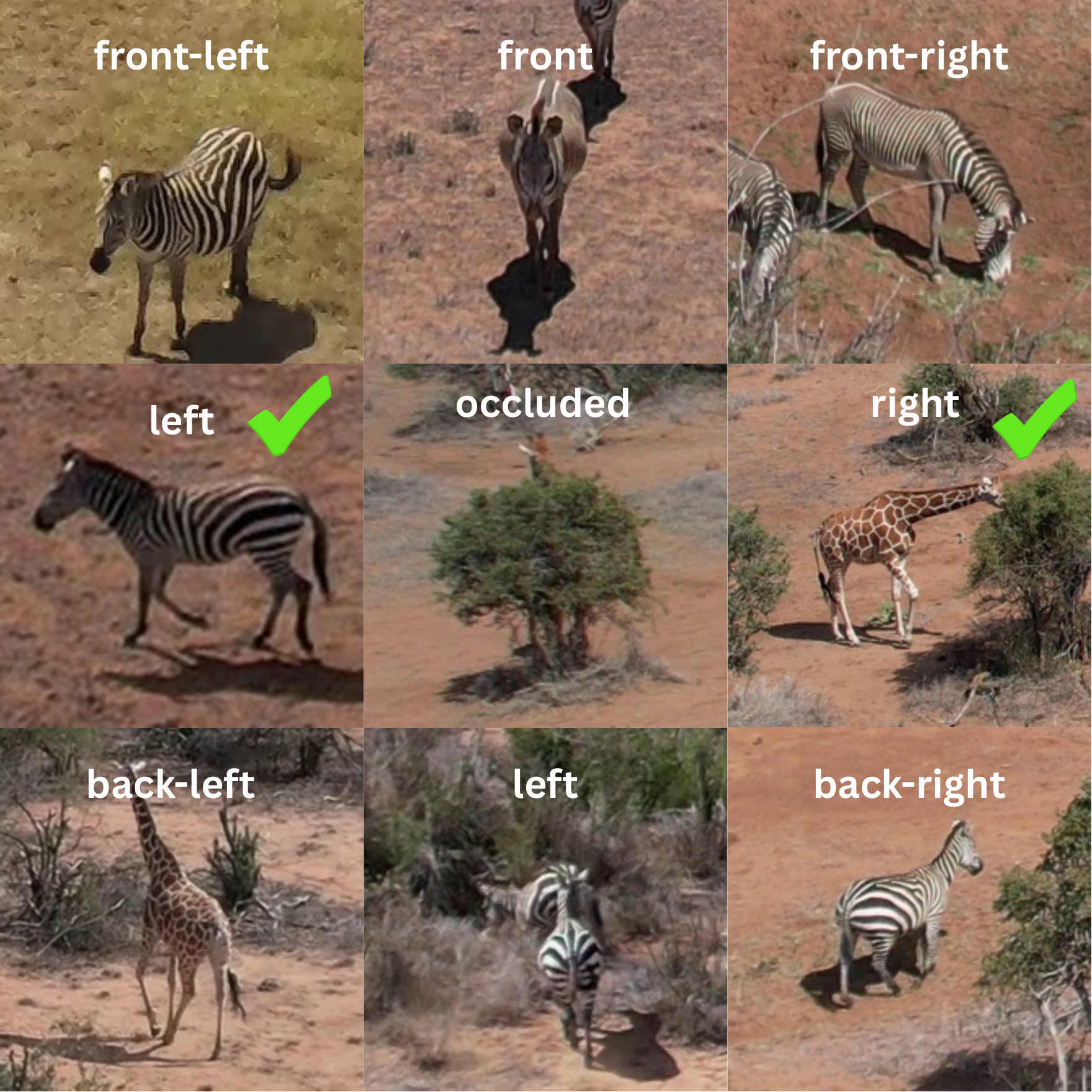}
    \caption{Example images from the pose training dataset. The left and right flanks are suitable for individual identification.}
    \label{fig:dataset}
\end{figure}


The detection and pose-classification models are built directly on the MMLA dataset \cite{kline2025mmla}.
MMLA is a collection of 155,074 frames of low-altitude aerial footage across six species (Plains zebra, Grevy's zebra, reticulated and Masai giraffe, Persian onager, African wild dog) from three field sites in Laikipia, Kenya and Ohio, United States. 
The YOLOv11m detector is the MMLA fine-tuned model, achieving 0.801 mAP50 across all species and 0.675 mAP50 on zebra specifically \cite{mmla_finetuned_yolo11m}. 
The DINOv2 pose classifier was fine-tuned on a manually curated subset of 1,023 MMLA aerial crops, sorted by hand into the 8 pose classes. 
Example images of the eight pose classes, plus occluded/negative examples, are illustrated in Fig. \ref{fig:dataset}.
DINOv2-small was chosen for its self-supervised pretraining on diverse imagery, which yields robust features that transfer to aerial animal crops, and for its small memory footprint, which is suitable for edge deployment \cite{oquab2023dinov2}. 
Training used label-aware horizontal flip augmentation bu swapping mirrored class pairs, color jitter, random crops, and weighted class sampling to handle natural pose imbalance in field footage.

At inference, the YOLO detection model has a latency of 4.7 ms on GPU, and DINOv2-small runs in approximately 3.5 ms per crop on GPU. Both models are well within the latency budget for real-time onboard control, which is typically 33ms for 30 frame-per-second video livestream \cite{Kline2025Edgenative}.
The navigation commands (yaw, forward velocity) are issued via SoftwarePilot \cite{angueira2023softwarepilot} to a Parrot Anafi platform running WildWing \cite{Kline2026WildWing}.

\section{Experimental Results}

We evaluate the proposed pipeline on drone footage collected in Kenya using both clipped evaluation sequences and full raw drone videos from the KABR collection.
The YOLOv11 detector achieved 0.801 mAP50 across all species and 0.675 mAP50 on zebra specifically \cite{mmla_finetuned_yolo11m}.
We evaluate the navigation pipeline on seven full, raw drone videos from the KABR collection that contain zebras \cite{KABR_Raw_Videos, kabr-mini-scene-videos}. Across these raw-video experiments, the system met the ratio-threshold capture condition in three videos, while the remaining videos resulted in timeouts or best-effort captures.
Results summarized in Table \ref{tab:raw_results}.

\begin{table}[h]
\centering
\caption{Evaluation results on full raw UAV videos from the KABR collection.}
\begin{tabular}{lccc}
\toprule
Video & Frame & Capture Type & BBox \\
\midrule
DJI\_0006 & 585 & timeout & 39$\times$43 \\
DJI\_0007 & 2005 & timeout & 133$\times$73 \\
DJI\_0018 & 123 & ratio & 400$\times$318 \\
DJI\_0019 & 6 & ratio & 573$\times$443 \\
DJI\_0070 & 764 & timeout & 158$\times$121 \\
DJI\_0142 & 305 & timeout & 128$\times$117 \\
DJI\_0207 & 6 & ratio & 441$\times$448 \\
\bottomrule
\end{tabular}
\label{tab:raw_results}
\end{table}

Across the seven full drone videos, the system reached the ratio-threshold capture condition in 3 of 7 videos, corresponding to a 42.9\% raw-video capture success rate. Successful captures occurred substantially earlier than timeout cases, with an average capture frame of 45.0 compared to 914.8 for timeout/best-effort cases. Successful captures also produced much larger bounding boxes on average (471.3$\times$403.0 pixels) than timeout cases (114.5$\times$88.5 pixels). Detection confidence remained high in both groups, averaging 0.83 for successful captures and 0.77 for timeout cases, suggesting that the main limitation was target scale and proximity rather than detector confidence.

These results demonstrate that pose-aware navigation can help obtain identification-quality wildlife images while also highlighting the practical challenges of autonomous visual acquisition in real-world field environments.

\section{Discussion}
Our preliminary experiments demonstrate that the DOAC pipeline substantially improves re-ID yield compared to passive capture. 
Detection alone produces a large volume of images, the majority of which are unsuitable for re-ID due to non-lateral poses or insufficient resolution at typical survey altitudes. 
The pose classifier achieves 75\% accuracy on held-out crops, and the integrated system successfully executes orient and approach maneuvers in field conditions.

Successful captures generally occur when the zebra already occupies a sufficiently large portion of the frame, producing bounding boxes up to 573$\times$443 pixels. In contrast, timeout cases produced substantially smaller detections, indicating that target proximity and image scale remain the primary bottlenecks for identification-quality image capture in unconstrained aerial footage.

Our results produce three consistent patterns.
First, pose filtering is critical: a large fraction of detections in uncontrolled field conditions present non-lateral views, and discarding these without maneuver would waste viable identification opportunities. However, the orient stage recovers a meaningful proportion of initially unsuitable detections. 
Second, the resolution threshold is binding: at standard survey altitudes (above 30m), most detections fall below the 500×500 pixel bounding box threshold, confirming that passive hovering is insufficient and active proximity control is necessary for re-ID-quality capture. 
Third, environmental factors remain a challenge: sparse detections in some sequences reflect vegetation occlusion and unfavorable lighting, consistent with findings from prior drone ecology work \cite{Corcoran2021Automated}. 
Robustness to these conditions is an important direction for future work.

\section{Conclusion and Future Work}
We present a proof-of-concept autonomous drone system for re-ID-optimized wildlife image collection. 
By treating image capture as an active perception problem, where the drone reasons about pose and resolution in real time, we move beyond the passive sensing paradigm that characterizes most drone-based wildlife monitoring. The preliminary results from our zebra case study demonstrate the feasibility of the detect–orient–approach–capture loop under real ecological conditions.

Our current results are preliminary and require additional testing. 
A rigorous quantitative comparison against expert-piloted baselines, and end-to-end re-ID accuracy measurements using collected images as input to a re-ID model (e.g., MegaDescriptor \cite{cermak2024Wildlife}), remain as immediate next steps. 
The system currently targets a single individual per flight segment.
Multi-animal coordination and herd-level coverage are important extensions. Animal welfare constraints, including minimizing disturbance, integrating vigilance monitoring as in Kline et al. \cite{Kline2025Edgenative}, must be addressed for field deployment at scale. 
Adverse weather, nighttime operation, and navigating complex safety regulations remain open challenges for drone-based wildlife monitoring \cite{maalouf2026sora}.

More broadly, this work contributes to an emerging paradigm of task-aware embodied AI for ecology: autonomous systems that are not merely deployed in the field, but are designed around the downstream computational requirements of the analysis they enable \cite{Tuia022Perspectives, Pringle2025Opportunities}. 
As re-ID models grow more capable, the data collection systems that feed them must grow correspondingly more intelligent.

\section*{Data and Code Availability Statement}
The \href{https://huggingface.co/datasets/imageomics/KABR-poses}{pose dataset} and fine-tuned DinoV2 \href{https://huggingface.co/imageomics/kabr-dino-pose}{model} are available on HuggingFace. The code is available on \href{https://github.com/Imageomics/individual-id-drones}{GitHub}.

\section*{Acknowledgments}
This project is supported by the U.S. National Science Foundation (NSF) \href{https://icicle.osu.edu/}{AI Institute for Intelligent Cyberinfrastructure with Computational Learning in the Environment (ICICLE)} under \href{https://www.nsf.gov/awardsearch/showAward?AWD_ID=2112606}{Award No. 2112606} 
and the \href{https://imageomics.org/}{Imageomics Institute} under \href{https://www.nsf.gov/awardsearch/showAward?AWD_ID=2118240}{Award No. 2118240} (Imageomics: A New Frontier of Biological Information Powered by Knowledge-Guided Machine Learning). 
{
    \small
    \bibliographystyle{ieeenat_fullname}
    \bibliography{main}

@article{Andrew2019Deep, series={IEEE International Conference on Intelligent Robots and Systems}, title={Deep Learning for Exploration and Recovery of Uncharted and Dynamic Targets from UAV-like Vision}, ISSN={9781538680957}, DOI={10.1109/IROS.2018.8593751}, journal={2018 IEEE/RSJ International Conference on Intelligent Robots and Systems (IROS 2018)}, publisher={Institute of Electrical and Electronics Engineers (IEEE)}, author={Andrew, William and Greatwood, Colin and Burghardt, Tilo}, year={2019}, month=jan, pages={1124–1131}, collection={IEEE International Conference on Intelligent Robots and Systems} }

@article{bergerwolf2017wildbook, title={Wildbook: Crowdsourcing, computer vision, and data science for conservation}, url={http://arxiv.org/abs/1710.08880}, DOI={10.48550/arXiv.1710.08880}, abstractNote={Photographs, taken by field scientists, tourists, automated cameras, and incidental photographers, are the most abundant source of data on wildlife today. Wildbook is an autonomous computational system that starts from massive collections of images and, by detecting various species of animals and identifying individuals, combined with sophisticated data management, turns them into high resolution information database, enabling scientific inquiry, conservation, and citizen science. We have built Wildbooks for whales (flukebook.org), sharks (whaleshark.org), two species of zebras (Grevy’s and plains), and several others. In January 2016, Wildbook enabled the first ever full species (the endangered Grevy’s zebra) census using photographs taken by ordinary citizens in Kenya. The resulting numbers are now the official species census used by IUCN Red List: http://www.iucnredlist.org/details/7950/0. In 2016, Wildbook partnered up with WWF to build Wildbook for Sea Turtles, Internet of Turtles (IoT), as well as systems for seals and lynx. Most recently, we have demonstrated that we can now use publicly available social media images to count and track wild animals. In this paper we present and discuss both the impact and challenges that the use of crowdsourced images can have on wildlife conservation.}, note={arXiv:1710.08880 [cs]}, number={arXiv:1710.08880}, publisher={arXiv}, author={Berger-Wolf, Tanya Y. and Rubenstein, Daniel I. and Stewart, Charles V. and Holmberg, Jason A. and Parham, Jason and Menon, Sreejith and Crall, Jonathan and Van Oast, Jon and Kiciman, Emre and Joppa, Lucas}, year={2017}, month=oct }

@inproceedings{cermak2024Wildlife, address={Waikoloa, HI, USA}, title={WildlifeDatasets: An open-source toolkit for animal re-identification}, rights={https://doi.org/10.15223/policy-029}, ISBN={9798350318920}, url={https://ieeexplore.ieee.org/document/10483925/}, DOI={10.1109/WACV57701.2024.00585}, abstractNote={In this paper, we present WildlifeDatasets – an opensource toolkit intended primarily for ecologists and computer-vision / machine-learning researchers. The WildlifeDatasets is written in Python, allows straightforward access to publicly available wildlife datasets, and provides a wide variety of methods for dataset pre-processing, performance analysis, and model fine-tuning. We showcase the toolkit in various scenarios and baseline experiments, including, to the best of our knowledge, the most comprehensive experimental comparison of datasets and methods for wildlife re-identification, including both local descriptors and deep learning approaches. Furthermore, we provide the first-ever foundation model for individual re-identification within a wide range of species – MegaDescriptor – that provides state-of-the-art performance on animal re-identification datasets and outperforms other pretrained models such as CLIP and DINOv2 by a significant margin. To make the model available to the general public and to allow easy integration with any existing wildlife monitoring applications, we provide multiple MegaDescriptor flavors (i.e., Small, Medium, and Large) through the HuggingFace hub.}, booktitle={2024 IEEE/CVF Winter Conference on Applications of Computer Vision (WACV)}, publisher={IEEE}, author={Čermák, Vojtěch and Picek, Lukas and Adam, Lukáš and Papafitsoros, Kostas}, year={2024}, month=jan, pages={5941–5951}, language={en} }

@article{Corcoran2021Automated, title={Automated detection of wildlife using drones: Synthesis, opportunities and constraints}, volume={12}, rights={© 2021 British Ecological Society}, ISSN={2041-210X}, DOI={10.1111/2041-210X.13581}, abstractNote={Accurate detection of individual animals is integral to the management of vulnerable wildlife species, but often difficult and costly to achieve for species that occur over wide or inaccessible areas or engage in cryptic behaviours. There is a growing acceptance of the use of drones (also known as unmanned aerial vehicles, UAVs and remotely piloted aircraft systems, RPAS) to detect wildlife, largely because of the capacity for drones to rapidly cover large areas compared to ground survey methods. While drones can aid the capture of large amounts of imagery, detection requires either manual evaluation of the imagery or automated detection using machine learning algorithms. While manual evaluation of drone-acquired imagery is possible and sometimes necessary, the powerful combination of drones with automated detection of wildlife in this imagery is much faster and, in some cases, more accurate than using human observers. Despite the great potential of this emerging approach, most attention to date has been paid to the development of algorithms, and little is known about the constraints around successful detection (P. W. J. Baxter, and G. Hamilton, 2018, Ecosphere, 9, e02194). We reviewed studies that were conducted over the last 5 years in which wildlife species were detected automatically in drone-acquired imagery to understand how technological constraints, environmental conditions and ecological traits of target species impact detection with automated methods. From this review, we found that automated detection could be achieved for a wider range of species and under a greater variety of environmental conditions than reported in previous reviews of automated and manual detection in drone-acquired imagery. A high probability of automated detection could be achieved efficiently using fixed-wing platforms and RGB sensors for species that were large and occurred in open and homogeneous environments with little vegetation or variation in topography while infrared sensors and multirotor platforms were necessary to successfully detect small, elusive species in complex habitats. The insight gained in this review could allow conservation managers to use drones and machine learning algorithms more accurately and efficiently to conduct abundance data on vulnerable populations that is critical to their conservation.}, number={6}, journal={Methods in Ecology and Evolution}, author={Corcoran, Evangeline and Winsen, Megan and Sudholz, Ashlee and Hamilton, Grant}, year={2021}, pages={1103–1114}, language={en} }

@article{Grushchak2023Decentralized, title={Decentralized Multi-Drone Coordination for Wildlife Video Acquisition}, abstractNote={One of the best sources of information for biologists and ethologists to study wildlife behavior is video footage; in particular, aerial video footage provides a unique perspective on the behavior of animals in their natural habitat. Numerous wildlife behavioral studies have demonstrated the effectiveness of UAVs for collecting video data of group-living animals. However, in contrast with well-established techniques for static video acquisition, the deployment of UAVs for wildlife video acquisition requires human operators to manually control and coordinate the drones while minimizing disturbance to animals. To scale UAVs missions to obtain sufﬁcient spatiotemporal resolution, reliance on manual operations is impractical. In this paper, we present a decentralized multi-drone coordination system for wildlife video acquisition using UAVs that leverages a novel k-coverage algorithm speciﬁcally designed to cover herds. In particular, it is based on a hierarchical clustering approach to ﬁnd the herds’ centroids, then it coordinates multiple drones in a decentralized fashion to cover them from multiple points of view. We introduce a set of metrics to evaluate the effectiveness of the proposed approach via simulation, ﬁnding that the proposed approach improves noticeably over the present state of the art.}, author={Grushchak, Denys}, language={en}, year = {2023}}

@inproceedings{Kholiavchenko2024KABR, title={KABR: In-Situ Dataset for Kenyan Animal Behavior Recognition From Drone Videos}, rights={All rights reserved}, url={https://openaccess.thecvf.com/content/WACV2024W/CV4Smalls/html/Kholiavchenko_KABR_In-Situ_Dataset_for_Kenyan_Animal_Behavior_Recognition_From_Drone_WACVW_2024_paper.html}, author={Kholiavchenko, Maksim and Kline, Jenna and Ramirez, Michelle and Stevens, Sam and Sheets, Alec and Babu, Reshma and Banerji, Namrata and Campolongo, Elizabeth and Thompson, Matthew and Van Tiel, Nina and Miliko, Jackson and Bessa, Eduardo and Duporge, Isla and Berger-Wolf, Tanya and Rubenstein, Daniel and Stewart, Charles}, year={2024}, pages={31–40}, language={en} }

@article{Kline2025Studying, title={Studying collective animal behaviour with drones and computer vision}, volume={n/a}, ISSN={2041-210X}, url={https://onlinelibrary.wiley.com/doi/abs/10.1111/2041-210X.70128}, DOI={10.1111/2041-210X.70128}, abstractNote={Drones are increasingly popular for collecting behaviour data of group-living animals, offering inexpensive and minimally disruptive observation methods. Imagery collected by drones can be rapidly analysed using computer vision techniques to extract information, including behaviour classification, habitat analysis and identification of individual animals. While computer vision techniques can rapidly analyse drone-collected data, the success of these analyses often depends on careful mission planning that considers downstream computational requirements—a critical factor frequently overlooked in current studies. We present a comprehensive summary of research in the growing AI-driven animal ecology (ADAE) field, which integrates data collection with automated computational analysis focused on aerial imagery for collective animal behaviour studies. We systematically analyse current methodologies, technical challenges and emerging solutions in this field, from drone mission planning to behavioural inference. We illustrate computer vision pipelines that infer behaviour from drone imagery and present the computer vision tasks used for each step. We map specific computational tasks to their ecological applications, providing a framework for future research design. Our analysis reveals AI-driven animal ecology studies for collective animal behaviour using drone imagery focus on detection and classification computer vision tasks. While convolutional neural networks (CNNs) remain dominant for detection and classification tasks, newer architectures like transformer-based models and specialized video analysis networks (e.g. X3D, I3D, SlowFast) designed for temporal pattern recognition are gaining traction for pose estimation and behaviour inference. However, reported model accuracy varies widely by computer vision task, species, habitats and evaluation metrics, complicating meaningful comparisons between studies. Based on current trends, we conclude semi-autonomous drone missions will be increasingly used to study collective animal behaviour. While manual drone operation remains prevalent, autonomous drone manoeuvrers, powered by edge AI, can scale and standardise collective animal behavioural studies while reducing the risk of disturbance and improving data quality. We propose guidelines for AI-driven animal ecology drone studies adaptable to various computer vision tasks, species and habitats. This approach aims to collect high-quality behaviour data while minimising disruption to the ecosystem.}, number={n/a}, journal={Methods in Ecology and Evolution}, author={Kline, Jenna and Afridi, Saadia and Rolland, Edouard G. A. and Maalouf, Guy and Laporte-Devylder, Lucie and Stewart, Christopher and Crofoot, Margaret and Stewart, Charles V. and Rubenstein, Daniel I. and Berger-Wolf, Tanya}, language={en}, year={2025} }

@article{Kline2024Integrating, title={Integrating Biological Data into Autonomous Remote Sensing Systems for In Situ Imageomics: A Case Study for Kenyan Animal Behavior Sensing with Unmanned Aerial Vehicles (UAVs)}, url={https://openreview.net/forum?id=HuK2e12l3a}, abstractNote={In situ imageomics leverages machine learning techniques to infer biological traits from images collected in the field, or in situ, to study individuals organisms, groups of wildlife, and whole ecosystems. In situ imageomics datasets provide real-time social and environmental context to inferred biological traits, which can enable new, data-driven conservation and ecosystem management. The development of machine learning techniques to extract biological traits from images are impeded by the volume and quality data required to train such models. Autonomous, unmanned aerial vehicles (UAVs), are well suited to collect in situ imageomics data as they can traverse remote terrain quickly to collect large volumes of data, with greater consistency and reliability compared to manually piloted missions. However, little guidance exists on optimizing autonomous UAV missions for the purposes of remote sensing for conservation and biodiversity monitoring. The UAV video dataset curated by KABR: In-Situ Dataset for Kenyan Animal Behavior Recognition from Drone Videos (Kholiavchenko et al. 2024) took three weeks to collect, a time-consuming and expensive to endeavor. Our analysis of KABR revealed that a third of the videos gathered were unusable for the purposes of inferring wildlife behavior. We analyzed the flight telemetry data from portions of UAV videos that were usable for inferring wildlife behavior, and demonstrate how these insights can be integrated into an autonomous remote sensing system to track in situ wildlife. Our autonomous remote sensing system optimizes the UAV’s actions to increase the yield of usable data, and matches the flight path of an expert pilot with an 87% accuracy rate, representing an 18.2% improvement in accuracy over previously proposed methods.}, journal={First Workshop on Imageomics}, author={Kline, Jenna and Berger-Wolf, Tanya and Kholiavchenko, Maksim and Brookes, Otto and Stewart, Charles V. and Stewart, Christopher}, year={2024} }

@inproceedings{Kline2025Edgenative, address={New York, NY, USA}, series={SEC ’25}, title={Edge-Native, Behavior-Adaptive Drone System for Wildlife Monitoring}, ISBN={9798400722387}, url={https://dl.acm.org/doi/10.1145/3769102.3774245}, DOI={10.1145/3769102.3774245}, abstractNote={Wildlife monitoring with drones must balance competing demands: approaching close enough to capture behaviorally-relevant video while avoiding stress responses that compromise animal welfare and data validity. Human operators face a fundamental attentional bottleneck: they cannot simultaneously control drone operations and monitor vigilance states across entire animal groups. By the time elevated vigilance becomes obvious, an adverse flee response by the animals may be unavoidable. To solve this challenge, we present an edge-native, behavior-adaptive drone system for wildlife monitoring. This configurable decision-support system augments operator expertise with automated group-level vigilance monitoring. Our system continuously tracks individual behaviors using YOLOv11m detection and YOLO-Behavior classification, aggregates vigilance states into a real-time group stress metric, and provides graduated alerts (alert vigilance → flee response) with operator-tunable thresholds for context-specific calibration. We derive service-level objectives (SLOs) from video frame rates and behavioral dynamics: to monitor 30fps video streams in real-time, our system must complete detection and classification within 33ms per frame. Our edge-native pipeline achieves 23.8ms total inference on GPU-accelerated hardware, meeting this constraint with a substantial margin. Retrospective analysis of seven wildlife monitoring missions demonstrates detection capability and quantifies the cost of reactive control: manual piloting results in 14 seconds average adverse behavior duration with 71.9% usable frames. Our analysis reveals operators could have received actionable alerts 51s before animals fled in 57% of missions. Simulating 5-second operator intervention yields a projected performance of 82.8% usable frames with 1-second adverse behavior duration, a 93% reduction compared to manual piloting.}, booktitle={Proceedings of the Tenth ACM/IEEE Symposium on Edge Computing}, publisher={Association for Computing Machinery}, author={Kline, Jenna and Katole, Rugved and Berger-Wolf, Tanya and Stewart, Christopher}, year={2025}, month=dec, pages={1–9}, collection={SEC ’25} }

@article{Kline2026WildWing, title={WildWing: An open-source, autonomous and affordable UAS for animal behaviour video monitoring}, volume={n/a}, ISSN={2041-210X}, url={https://onlinelibrary.wiley.com/doi/abs/10.1111/2041-210X.70018}, DOI={10.1111/2041-210X.70018}, abstractNote={Drones have become invaluable tools for studying animal behaviour in the wild, enabling researchers to collect aerial video data of group-living animals. However, manually piloting drones to track animal groups consistently is challenging due to complex factors such as terrain, vegetation, group spread and movement patterns. The variability in manual piloting can result in unusable data for downstream behavioural analysis, making it difficult to collect standardized datasets for studying collective animal behaviour. To address these challenges, we present WildWing, a complete hardware and software open-source unmanned aerial system (UAS) for autonomously collecting behavioural video data of group-living animals. The system’s main goal is to automate and standardize the collection of high-quality aerial footage suitable for computer vision-based behaviour analysis. We provide a novel navigation policy to autonomously track animal groups while maintaining optimal camera angles and distances for behavioural analysis, reducing the inconsistencies inherent in manual piloting. The complete WildWing system costs only $650 and incorporates drone hardware with custom software that integrates ecological knowledge into autonomous navigation decisions. The system produces 4 K resolution video at 30 fps while automatically maintaining appropriate distances and angles for behaviour analysis. We validate the system through field deployments tracking groups of Grevy’s zebras, giraffes and Przewalski’s horses at The Wilds conservation centre, demonstrating its ability to collect usable behavioural data consistently. By automating the data collection process, WildWing helps ensure consistent, high-quality video data suitable for computer vision analysis of animal behaviour. This standardization is crucial for developing robust automated behaviour recognition systems to help researchers study and monitor wildlife populations at scale. The open-source nature of WildWing makes autonomous behavioural data collection more accessible to researchers, enabling wider application of drone-based behavioural monitoring in conservation and ecological research.}, number={n/a}, journal={Methods in Ecology and Evolution}, author={Kline, Jenna and Zhong, Alison and Irizarry, Kevyn and Stewart, Charles V. and Stewart, Christopher and Rubenstein, Daniel I. and Berger-Wolf, Tanya}, language={en}, year={2026} }

@article{khanam2024yolov11,
  title={Yolov11: An overview of the key architectural enhancements},
  author={Khanam, Rahima and Hussain, Muhammad},
  journal={arXiv preprint arXiv:2410.17725},
  year={2024}
}
}


\end{document}